\documentclass{ecai} 

\usepackage{algorithm}
\usepackage{algorithmic}
\usepackage{amssymb}
\usepackage{amsmath}
\usepackage{amsthm}
\usepackage{booktabs}
\usepackage{color}
\usepackage{enumitem}
\usepackage{graphicx}
\usepackage{hyperref}
\usepackage{latexsym}
\usepackage{makecell}

\newcommand{\Sc}{\mathcal{S}}
\newcommand{\Ac}{\mathcal{A}}
\newcommand{\Tc}{\mathcal{T}}

\newcommand{\Rc}{\mathcal{R}}

\newcommand{\BibTeX}{B\kern-.05em{\sc i\kern-.025em b}\kern-.08em\TeX}

\begin{document}

%%%%%%%%%%%%%%%%%%%%%%%%%%%%%%%%%%%%%%%%%%%%%%%%%%%%%%%%%%%%%%%%%%%%%%%%

\begin{frontmatter}

%%% Use this command to specify your submission number.
%%% In doubleblind mode, it will be printed on the first page.

\paperid{4294} 

%%% Use this command to specify the title of your paper.

\title{Skill-Enhanced Reinforcement Learning Acceleration \\
from Heterogeneous Demonstrations}

%%% Use this combinations of commands to specify all authors of your 
%%% paper. Use \fnms{} and \snm{} to indicate everyone's first names 
%%% and surname. This will help the publisher with indexing the 
%%% proceedings. Please use a reasonable approximation in case your 
%%% name does not neatly split into "first names" and "surname".
%%% Specifying your ORCID digital identifier is optional. 
%%% Use the \thanks{} command to indicate one or more corresponding 
%%% authors and their email address(es). If so desired, you can specify
%%% author contributions using the \footnote{} command.

\author[$^*$]{\fnms{Hanping}~\snm{Zhang}}%\footnote{Our code is available at \href{https://github.com/jajajag/serla}{https://github.com/jajajag/serla}.}}
\author[$^*\dagger$]{\fnms{Yuhong}~\snm{Guo}}%\orcid{....-....-....-....}\footnotemark}

\address[$^*$]{School of Computer Science, Carleton University, Ottawa, Canada}
\address[$\dagger$]{Canada CIFAR AI Chair, Amii, Canada}
\address[]{\{jagzhang@cmail., yuhong.guo@\}carleton.ca}

%%% Use this environment to include an abstract of your paper.

\begin{abstract}
Learning from Demonstration (LfD)
is a well-established problem in Reinforcement Learning (RL), which 
aims to facilitate rapid RL 
by leveraging expert demonstrations to pre-train the RL agent. 
However, the limited availability of expert demonstration data often hinders 
its ability to effectively aid downstream RL learning.
To address this problem, we propose a novel two-stage method dubbed as Skill-enhanced Reinforcement 
Learning Acceleration (SeRLA). 
SeRLA introduces a skill-level adversarial Positive-Unlabeled (PU) learning model 
that extracts useful skill prior knowledge by learning from both expert demonstrations 
and general low-cost demonstrations in the offline prior learning stage. 
Building on this, it employs a skill-based soft actor-critic algorithm to 
leverage the acquired priors for efficient training of a skill policy network in the downstream online RL stage. 
In addition, we propose a simple skill-level data enhancement technique to 
mitigate data sparsity and further improve both skill prior learning and skill policy training. 
Experiments across multiple standard RL benchmarks demonstrate that 
SeRLA achieves state-of-the-art performance in accelerating reinforcement learning on downstream tasks, 
particularly in the early training phase.
\end{abstract}

\end{frontmatter}

\section{Introduction}

Despite the wide applicability of Reinforcement Learning (RL) across robotics~\cite{kober2013reinforcement}, video games~\cite{vinyals2017starcraft,berner2019dota}, and large language models~\cite{stiennon2020learning,ouyang2022training}, a conventional deep RL agent often requires numerous iterative interactions with the environment to learn a useful policy by maximizing the expected discounted cumulative reward~\cite{sutton2018reinforcement}, resulting in prolonged training periods and limited computational efficiency. This challenge becomes more pronounced in complex environments, where exploration is both costly and time-consuming. To overcome this problem, Learning from Demonstration (LfD), also known as imitation learning (IL), has been investigated to accelerate RL.
In LfD, the agent is pre-trained on a small offline demonstration dataset provided by human experts~\cite{argall2009survey,brys2015reinforcement} to acquire knowledge and learn behaviors that can be executed in the environment, 
which can then be leveraged to accelerate the online learning process of downstream RL tasks with fewer environment interactions.
Due to the limited availability of expert demonstration data, some recent studies seek to supplement the expert data with a large task-agnostic demonstration dataset collected inexpensively using methods such as autonomous exploration~\cite{hausman2018learning,Sharma2020Dynamics-Aware} or human-teleoperation~\cite{gupta2019relay,mandlekar2018roboturk}.

Skill-based RL, which acquires reusable skills---high-level behaviors composed of primitive actions---from expert demonstrations~\cite{lee2019composing,dalal2021accelerating}
or environment interactions~\cite{lee2020learning,eysenbach2018diversity,dalal2021accelerating,hausman2018learning}  
to guide reinforcement learning, shows great potential for advancing LfD.
Recently, researchers have introduced skill-based RL to LfD by learning reusable skill behaviors 
from demonstration data and deploying them for downstream tasks
\cite{gupta2019relay,pertsch2021accelerating,shi2022skillbased,hakhamaneshi2022hierarchical,xu2022aspire}. 
However, these previous studies 
either focus solely on learning from expert datasets
or treat general demonstration data as negative examples,
thereby impeding the effective utilization of low-cost demonstrations
that are widely available and 
may contain valuable fragmented skills.

In this paper, we propose a novel SeRLA method, which stands for \textbf{S}kill-\textbf{e}nhanced \textbf{R}einforcement \textbf{L}earning \textbf{A}cceleration from heterogeneous demonstrations, 
to address the problem of 
learning from heterogeneous demonstration data
and accelerating downstream RL with the learned knowledge. 
SeRLA accelerates RL by pursuing {\em skill-level} learning in two stages 
with three coherent components: 
a skill-level adversarial PU learning module, 
a skill-based soft actor-critic policy learning algorithm, 
and a skill-level data enhancement technique.
In the offline skill prior training stage, 
the skill-level adversarial PU learning module learns useful skill prior knowledge
by exploiting the general, task-agnostic demonstration data as unlabeled examples
in addition to the {\em positive} expert data, 
instead of simply differentiating them. 
This strategy facilitates improved utilization of the extensive low-cost demonstration data 
and can help alleviate the scarcity of the expert data.
In the online downstream RL policy training stage, 
a skill-based soft actor-critic algorithm 
is deployed to integrate skills learned in the offline stage 
and accelerate skill policy learning. 
Moreover, a simple but novel 
Skill-level Data Enhancement (SDE) 
technique is introduced to improve 
the robustness of skill learning and adaptation at both stages. 
We conduct experiments on four standard RL environments by comparing the proposed SeRLA
with the state-of-the-art skill-based imitation learning methods: 
SPiRL~\cite{pertsch2021accelerating} and model-based SkiMo~\cite{shi2022skillbased}.
The main contributions of this paper are summarized as follows:
\begin{itemize}
\item 
This is the first work that conducts skill-level Positive-Unlabeled Learning for LfD/IL. 
The proposed SeRLA 
uses low-cost, general demonstration data as unlabeled examples 
to statistically support skill learning from limited positive examples---i.e., 
expert demonstrations---through 
skill-level adversarial PU learning.
\item 
We propose a simple but novel skill-level data enhancement (SDE) technique, 
which automatically augments the skill-level representations 
for both the skill prior learning and the downstream policy learning processes  
to improve the robustness of the learned skill prior and 
accelerate the skill-policy function training. 
\item 
The proposed SeRLA produces effective empirical results
for accelerating downstream RL tasks. 
It largely outperforms the standard skill prior learning method, SPiRL,
while producing notable improvements over the state-of-the-art model-based skill-level method, 
SkiMo, in the early downstream training stage. 
\end{itemize}

%%%%%%%%%%%%%%%%%%%%%%%%%%%%%%%%%%%%%%%%%%%%%%%%%%%%%%%%%%%%%%%%%%%%%%%%%%%%%%%

\section{Related Works}

\subsection{Reinforcement Learning from Demonstrations}
Learning from Demonstration (LfD) or imitation learning (IL) 
aims to accelerate reinforcement learning for downstream RL tasks
by pre-training the RL agent 
on a small expert demonstration dataset typically without reward signals 
\cite{argall2009survey,brys2015reinforcement}. 
Unlike offline RL, which aims to learn optimal policies solely from offline data, 
LfD/IL focuses on accelerating downstream tasks.
In addition to the limited expert demonstrations in the form of 
a sequence of state-action pairs $\{(s_0,a_0),...,(s_t, a_t)\}$,
large task-agnostic demonstration datasets can also be collected inexpensively 
\cite{hausman2018learning,Sharma2020Dynamics-Aware,gupta2019relay,mandlekar2018roboturk} 
from the environment for extracting potential learnable behaviors through LfD. 
Apart from these existing works, recent works for LfD/IL also include
Behavior Cloning (BC)~\cite{argall2009survey}, Inverse Reinforcement Learning (IRL)~\cite{abbeel2004apprenticeship}, 
and Generative Adversarial Imitation Learning (GAIL)~\cite{ho2016generative}. 
BC enables the RL agent to learn a direct mapping between observations and corresponding actions from the demonstration dataset as a supervised learning problem. 
This method however has limited generalization ability and suffers from distribution shifts~\cite{ross2010efficient,ross2011reduction}. 
IRL infers reward functions from the demonstration data and trains the RL agent using standard RL algorithms~\cite{li2023accelerating}. 
Although IRL can transform imitation learning to a standard RL problem, 
it is computationally expensive and relies heavily on the effectiveness of the reward model. 
GAIL treats IL as a two-player zero-sum game with a generative adversarial network~\cite{goodfellow2020generative}, 
where a discriminator is trained to distinguish the behavior between the agent policy and the expert policy 
learned from the demonstrations. 
It solves the zero-sum game using minimax optimization, 
yielding similar behaviors between the agent policy and the expert policy. 
Despite the requirement of numerous interactions with the environment, GAIL demonstrates remarkable performance on IL.

\subsection{Skill-Based Reinforcement Learning}
As a popular approach to leveraging prior knowledge, 
skill-based RL methods 
extract reusable skills as abstracted long-horizon behavior sequences of actions~\cite{lee2019composing,lee2020learning,pertsch2021accelerating,lee2021adversarial,dalal2021accelerating,hausman2018learning,gupta2019relay,eysenbach2018diversity}. 
These skills can either be predefined by experts or extracted from online or offline datasets,
naturally supporting LfD or IL. 
One recent work 
\cite{pertsch2021accelerating} in this line 
introduces Skill-Prior RL (SPiRL), a hierarchical skill-based approach, 
to accelerate downstream RL tasks using learned skill priors from offline demonstration data. 
Another work~\cite{pertsch2021demonstrationguided} proposes 
a Skill-based Learning with Demonstration (SkiLD) approach, 
which uses a skill posterior to regularize the policy learning in the downstream task. 
In addition, 
the approach of Few-shot Imitation with Skill Transition Models (FIST)
\cite{hakhamaneshi2022hierarchical} 
learns skills from few-shot demonstration data and generalizes to unseen tasks. 
The Adaptive Skill Prior for RL (ASPiRe), introduced by \citet{xu2022aspire}, 
adaptively learns distinct skill priors from different datasets with specific weights. 
Furthermore, \citet{shi2022skillbased} developed a Skill-based Model-based RL framework (SkiMo) that applies planning on downstream tasks 
by using a skill dynamics model to select the optimal skill from different learned skills.
\citet{celik2024acquiring} introduced energy-based models 
to learn diverse skills via a mixture of experts.

\subsection{Positive-Unlabeled (PU) Learning}
Unlike the standard binary classifier that learns from positive and negative examples, 
PU learning utilizes only positive and unlabeled data~\cite{bekker2020learning},
where an unlabeled example can belong to either the positive or negative class. 
PU learning has demonstrated values in real-world applications with PU data 
such as medical diagnosis~\cite{claesen2015building}
and knowledge base construction~\cite{galarraga2015fast,zupanc2018estimating}. 
Prior works on PU learning focus on the loss functions and optimizers~\cite{du2014analysis,patrini2016loss}. 
A recent work by \citet{kiryo2017positive} builds large scale PU learning upon deep neural networks. 
\citet{zhao2024boosting} proposed a novel boosting framework aimed at improving PU classifier training efficiency.
More recently, \citet{xu2021positive} demonstrated the utility of PU Learning in imitation reward learning 
by replacing the adversarial loss with a PU loss in GAIL~\cite{ho2016generative} 
and developing a specific PU learning-based reward function to train the RL agent 
with expert demonstration data.
By contrast, our proposed work delivers a skill-level two-stage training framework
that can learn skill priors simultaneously from both limited expert data and low-cost general demonstration data
via a PU learning module,
while accelerating the downstream RL tasks.

%%%%%%%%%%%%%%%%%%%%%%%%%%%%%%%%%%%%%%%%%%%%%%%%%%%%%%%%%%%%%%%%%%%%%%%%%%%%%%%

\begin{figure*}[t]
\centering
\setlength{\abovecaptionskip}{0.cm}
\includegraphics[width=0.95\textwidth]{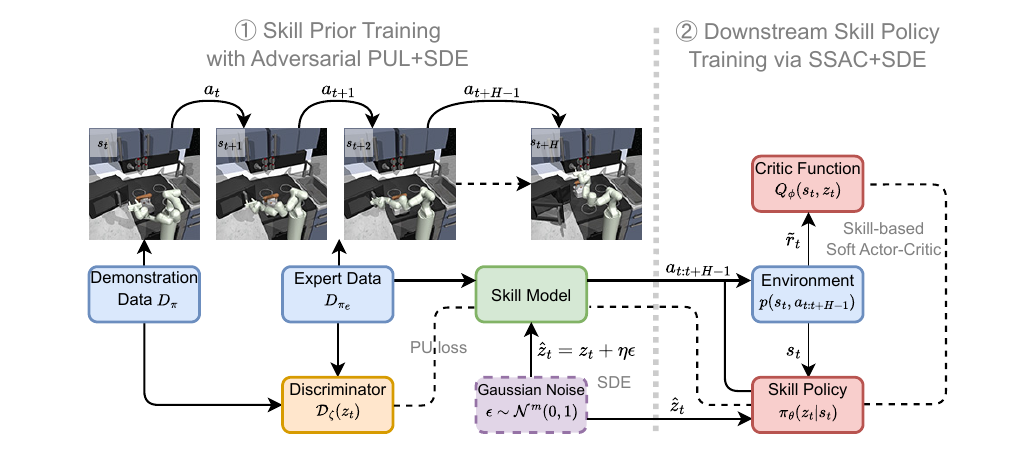}
\vskip .05in
\caption{The proposed method, SeRLA, pursues skill-level training in two stages: 
    offline skill prior training via skill-level PU Learning 
	and online downstream skill policy training via SSAC (Skill-based Soft Actor-Critic).
    \textbf{Left}: 
    Skill-level PU-Learning 
	incorporates a large random demonstration dataset $D_\pi$ 
	into skill learning on a small expert dataset $D_{\pi_e}$ 
	by adding a discriminator $\mathcal{D}_\zeta$ through an adversarial PU loss.
\textbf{Right}: 
The learned skill knowledge is exploited to accelerate the downstream online RL task 
through a model-free SSAC algorithm, which utilizes the 
	prior skill knowledge through behavior cloning. 
{\bf Moreover}, Skill-level Data Enhancement (SDE) 
	is proposed to further alleviate data sparsity and improve learning robustness 
	by employing skill augmentation in both the offline prior learning stage and the online downstream RL training stage.
}
\vskip 0.15in
\label{fig:architecture}
\end{figure*}

%%%%%%%%%%%%%%%%%%%%%%%%%%%%%%%%%%%%%%%%%%%%%%%%%%%%%%%%%%%%%%%%%%%%%%%%%%%%%%%
\section{Problem Setting}
Reinforcement Learning from Demonstrations (LfD) aims to accelerate the online downstream 
RL procedure by leveraging offline demonstration datasets. 
Specifically, we assume LfD has access to two demonstration datasets: 
a limited expert dataset $D_{\pi_e}$ and a low-cost general demonstration dataset $D_\pi$.
The expert dataset is a task-specific small offline dataset that contains expert demonstration trajectories
(state-action sequences)
$D_{\pi_e}=\{s_0, a_0, \cdots, s_t, a_t, \cdots\}$, which are generated by human experts or fully trained RL agents. 
The general demonstration dataset is a much larger 
task-agnostic offline dataset that consists of randomly collected trajectories 
$D_\pi=\{s_0, a_0, \cdots, s_t, a_t, \cdots\}$.
The action sequences contained in $D_{\pi_e}$ and $D_{\pi}$ 
can be denoted as $A_{\pi_e}$ and $A_\pi$, respectively.
While the general demonstration dataset does not provide as much useful information as the expert dataset, 
it may still contain short-horizon behaviors that, if properly extracted, 
can guide the RL agent to behave with feasible actions
and propel policy learning. 

The downstream RL task is a standard reinforcement learning problem that can be represented as a Markov Decision Process (MDP) 
$M=(\Sc, \Ac, \Tc, \Rc, \gamma)$, as described in~\cite{sutton2018reinforcement}. In this MDP, $\Sc$ is the state space, $\Ac$ is the action space, $\Tc: \Sc\times \Ac\to \Sc$ is the transition dynamics $p(s_{t+1}|s_t,a_t)$, $\Rc: \Sc\times \Ac\to \mathbb{R}$ is the reward function, and $\gamma\in (0, 1)$ is the discount factor. The goal is to learn an optimal policy $\pi^\star: \Sc \to \Ac$ that maximizes the expected discounted cumulative reward (return):
$\pi^\star=\arg\max_\pi\; J_r(\pi)=\mathbb{E}_{\pi}[\sum_{t=0}^{T} \gamma^t r_t]$.
The goal of this work is to 
learn useful skill prior from the offline heterogeneous demonstration datasets
and deploy such skill-level knowledge to facilitate fast policy training for the downstream RL task.

%%%%%%%%%%%%%%%%%%%%%%%%%%%%%%%%%%%%%%%%%%%%%%%%%%%%%%%%%%%%%%%%%%%%%%%%%%%%%%%
\section{Method}

The main architecture of the proposed SeRLA method is illustrated in Figure~\ref{fig:architecture}. 
It has two stages: the offline skill prior training stage with skill-level adversarial PU learning
and the online skill-based downstream policy training. 
The skill prior training induces useful high-level skill knowledge in form of skill priors
from the given heterogeneous demonstration datasets ($D_{\pi_e}$ and $D_\pi$),
which is then used to accelerate the downstream skill-based policy training
through a Skill-based Soft Actor-Critic (SSAC) algorithm.
Moreover, 
a simple skill-level data enhancement technique is further devised
for the two training stages to improve the overall performance. 
Below, we elaborate on these approach components. 

%%%%%%%%%%%%%%%%%%%%%%%%%%%%%%%%%%%%%%%%%%%%%%%%%%%%%%%%%%%%%%%%%%%%%%%%%%%%%%%
\subsection{Skill Prior Training with Adversarial PU Learning}
\label{section:spt}

In the skill prior training stage, 
we build a regularized deep autoencoder module with skill-level adversarial PU learning
to learn a conditional skill prior distribution function 
$q_\psi(z_t|s_t)$
from the trajectories provided in the two demonstration datasets,
where latent variables $\{z_t\}$ 
are used to capture the high level representations
of skills, each of which can be interpreted as an action sequence. 
The model includes a skill encoder network $q_\mu(\cdot)$, a skill decoder network $p_\nu(\cdot)$,
a skill prior network $q_\psi(\cdot)$, and a discriminator $\mathcal{D}_\zeta$.
The first three components can be learned 
from the expert data $D_{\pi_e}$ within a conventional autoencoder framework,
while the discriminator 
is innovatively deployed to incorporate the general demonstration data 
$D_{\pi}$ and alleviate 
skill data sparsity via adversarial PU learning.  

%%%%%%%%%%%%%%%%%%%%%%%%%%%%%%%%%%%%%%%%%%%%%%%%
\paragraph{Conventional Skill Learning Framework}
Under a deep autoencoder framework~\cite{pertsch2021accelerating},
the skill encoder $q_\mu(z_t|{\bf a}_t)$ takes an action sequence 
${\bf a}_t=\{a_t, ..., a_{t+H-1}\}$ with length $H$
as input and maps it to a latent skill embedding $z_t$. 
Conversely, the skill decoder $p_\nu(\hat{\bf a}_t|z_t)$ reconstructs
an action sequence $\hat{\textbf{a}}_t$ 
from a given skill $z_t$.
The autoencoder can be learned by minimizing 
a \textit{reconstruction loss} on 
the observed action sequences $A_{\pi_e}$ 
from the expert demonstrations $D_{\pi_e}$:
\begin{align}
\label{equation:rec}
	\!\!\! L_{rec}(\nu,\mu)=
\mathbb{E}_{{\bf a}_t\sim A_{\pi_e}}
	\ell_{ls}\big(\hat{\bf a}_t\sim p_\nu\big(\cdot| 
	z_t\!\sim\! q_\mu(\!\cdot|{\bf a}_t)\big), {\bf a}_t \big),
\end{align}
where $\ell_{ls}(\cdot,\cdot)$ is the standard least squares loss. 
The skill prior network $q_\psi(z_t|s_t)$ generates a skill $z_t$ 
from a given starting state $s_t$.
It is designed to support policy network training for the downstream tasks 
by encoding the expert behavior in a given state in terms of skills
from the expert dataset. 
Ideally, given a pair of observed state and action sequence $(s_t, {\bf a}_t)$,
the skills produced by the encoder $q_\mu(z_t|{\bf a}_t)$ and the prior network $q_\psi(z_t|s_t)$
should be consistent.  
Hence, the skill prior network can be learned together with the encoder
by minimizing the following \textit{prior training loss}:
\begin{align}
\!\!	L_{prior}(\psi,\mu)=
\mathbb{E}_{(s_t, {\bf a}_t) \sim D_{\pi_e}} 
	\mathcal{L}_{KL}(q_\mu(z_t|{\bf a}_t), q_\psi(z_t|s_t)),
\end{align}
where $\mathcal{L}_{KL}(\cdot,\cdot)$ denotes the Kullback Leibler (KL) divergence function. 
Moreover, 
a standard Gaussian distribution prior $p(z_t)$ $=$ $\mathcal{N}(0, 1)$ 
can be deployed 
to regularize the skill embedding space
with the following \textit{regularization loss}: 
\begin{align}
	L_{reg}(\mu)=\mathbb{E}_{{\bf a}_t\sim A_{\pi_e}}\mathcal{L}_{KL}(q_\mu(z_t|{\bf a}_t), p(z_t))
\end{align}
%
%%%%%%%%%%%%%%%%%%%%%%%%%%%%%%%%%%%%%%%%%%%%%%%%%%%
\subsubsection{Skill-level Adversarial PU Learning}

Different from the expert data, the randomly collected low-cost, large demonstration dataset $D_{\pi}$
can present a great number of short-horizon behaviors,
some of which can be meaningful while many others can be spontaneous or arbitrary.
Hence it is not suitable to directly deploy $D_{\pi}$ in the autoencoder model above in the same way as 
the expert data $D_{\pi_e}$.
To effectively filter out the noisy behaviors and exploit the useful ones from $D_{\pi}$,
we deploy a PU learning scheme to 
perform skill learning simultaneously from both the small expert data $D_{\pi_e}$
and the large general demonstration data $D_{\pi}$.
PU learning is a variant of supervised learning that learns a binary classifier 
from positive and unlabeled data~\cite{bekker2020learning}.
By modeling the unlabeled data instead of assuming it is entirely negative, 
PU learning can exploit additional information and reduce bias compared to naive methods.
Specifically, we treat the skills (capturing the behaviors) from the expert data, 
$Z_e =q_\mu(A_{\pi_e})$, 
as {\em positive examples}
(i.e., useful skills),
and treat skills from the general demonstration data, 
$Z = q_\mu(A_\pi)$,
as {\em unlabeled examples}
that can be either positive or negative.  
Then we propose to learn a binary probabilistic discriminator $\mathcal{D}_\zeta$ 
from the positive and unlabeled skill examples 
by adapting a standard non-negative PU risk 
function derived in the literature~\cite{kiryo2017positive}
into the skill-level learning: 
\begin{align}
	{L}_{\mathcal{D}_\zeta}^{pu}(q_\mu(A_{\pi_e}), &q_\mu(A_\pi)) = 
	\lambda {L}_{\mathcal{D}_\zeta}^1(q_\mu(A_{\pi_e})) +\nonumber\\ 
& \max(
    -\xi, {L}_{\mathcal{D}_\zeta}^0(q_\mu(A_\pi)) - \lambda {L}_{\mathcal{D}_\zeta}^0(q_\mu(A_{\pi_e}))) 
\label{equation:pu_loss}
\end{align}
where $\lambda > 0$ and $\xi\geq 0$ are hyperparameters. 
Here the true positive risk ${L}_{\mathcal{D}_\zeta}^1(q_\mu(A_{\pi_e}))$ 
is calculated on positive skill examples $Z_e =q_\mu(A_{\pi_e})$, 
while the true negative risk is calculated on both positive and unlabeled data
($Z_e$ and $Z$)
using two terms,  
${L}_{\mathcal{D}_\zeta}^0(q_\mu(A_{\pi_e}))$
and ${L}_{\mathcal{D}_\zeta}^0(q_\mu(A_{\pi}))$. 
These risk terms are defined in terms of the discriminator $\mathcal{D}_\zeta$ as follows:
\begin{align}
	{L}_{\mathcal{D}_\zeta}^1(q_\mu(A_{\pi_e})) &= \mathop{\mathbb{E}}_{{\bf a}_t\sim A_{\pi_e}}
	[\log(1-\mathcal{D}_\zeta(z_t\!\sim\! q_\mu(\cdot|{\bf a}_t)))]\\
	{L}_{\mathcal{D}_\zeta}^0(q_\mu(A_\pi)) &= \mathop{\mathbb{E}}_{{\bf a}_t\sim A_\pi} [\log(\mathcal{D}_\zeta(z_t\sim q_\mu(\cdot|{\bf a}_t)))]\\
	{L}_{\mathcal{D}_\zeta}^0(q_\mu(A_{\pi_e})) &= \mathop{\mathbb{E}}_{{\bf a}_t\sim A_{\pi_e}} [\log(\mathcal{D}_\zeta(z_t\sim q_\mu(\cdot|{\bf a}_t)))]
\end{align}
where $\mathcal{D}_\zeta(z_t)$ predicts the probability of the given
skill vector $z_t$ being a positive example 
and $(1-\mathcal{D}_\zeta(z_t))$ denotes the probability of the given
skill vector $z_t$ being a negative example. 

This PU loss ${L}_{\mathcal{D}_\zeta}^{pu}$
can be integrated into the deep skill learning model in an {\em adversarial} manner
to enable the exploitation of the large demonstration data $D_{\pi}$: 
the discriminator $\mathcal{D}_\zeta$ will be learned 
to {\em minimize} the PU loss in Eq.(\ref{equation:pu_loss}) given the skill examples extracted,
while the skill encoder $q_\mu(\cdot)$ will be learned to {\em maximize} the PU loss,
aiming to alleviate the scarcity of expert data and 
generalize the skill learning to the large general demonstration data $D_{\pi}$. 

%%%%%%%%%%%%%%%%%%%%%%%%%%%%%%%%%%%%%%%%%%%%%%%%%%%%%%%%%%%%%%%%%%%%%%%%%%%%%%%
\begin{algorithm}[t]
\caption{Skill Prior Training via PU Learning}
\label{algorithm:spt}
\textbf{Input:} 
Expert data $D_{\pi_e}$,
general demonstration data $D_\pi$ \\
\textbf{Initialize:}
	Encoder $q_\mu(\cdot)$, decoder $p_\nu(\cdot)$, skill prior $q_\psi(\cdot)$, and discriminator $\mathcal{D}_\zeta(\cdot)$ \\
\textbf{Output:} Trained skill prior $q_\psi(z_t|s_t)$, and decoder $p_\nu(a_{t:t+H-1}|z_t)$\\
\textbf{Procedure:}
\begin{algorithmic}[1]
\FOR{each iteration}
\FOR{every $H$ environment steps}
	\STATE Sample $s_t$ and sequence ${\bf a}_t=a_{t:t+H-1}$ from $D_{\pi_e}$ 
	\STATE Sample action sequence ${\bf a}_t'=a_{t':t'\!+\!H\!-\!1}'$ from $D_\pi$ (or $A_{\pi}$) 
	\STATE Update $\mu$, $\nu$, $\psi$ by minimizing Eq.(\ref{equation:joint_loss})
	\STATE Update $\zeta$ by maximizing Eq.(\ref{equation:joint_loss})
\ENDFOR
\ENDFOR
\end{algorithmic}
\end{algorithm}
%

%%%%%%%%%%%%%%%%%%%%%%%%%%%%
\paragraph{Skill Prior Training Algorithm}
The total loss function for adversarial PU-learning based 
skill prior training can be expressed as the sum of four terms: 
\begin{align}
L(\mu, \nu, \psi)=&L_{rec}(\nu,\mu) + L_{prior}(\psi,\mu) + \beta L_{reg}(\mu)\nonumber\\
	&- \rho \min\nolimits_{\zeta} {L}_{\mathcal{D}_\zeta}^{pu}(q_\mu(A_{\pi_e}), q_\mu(A_\pi)) 
\label{equation:joint_loss}
\end{align}
where $\beta$ and $\rho$ are tradeoff hyperparameters;
the reconstruction loss $L_{rec}$ 
enforces consistency between the skill embedding $z_t$ and the action sequence 
${\bf a}_t$;
the prior training loss $L_{prior}$ ensures that the generated skill is consistent with the current state and action sequence; 
$L_{reg}$ regularizes the skill embedding space;
and the PU loss ${L}_{D_\zeta}^{pu}$ is used to effectively incorporate large random demonstration data
into skill learning in an adversarial manner. 
The joint training of all these components is 
expected to effectively and proficiently learn valuable skill knowledge 
from the unified heterogeneous offline demonstration datasets. 

Algorithm~\ref{algorithm:spt} outlines the main steps of stochastic skill prior training. 
At each timestep $t$, we collect the current state $s_t$ and an action sequence $a_{t:t+H-1}$ from the expert dataset $D_{\pi_e}$ in an $H$-step rollout. 
Similarly we collect a sequence of actions $a_{t':t'+H-1}$ from the general demonstration dataset. 
We jointly learn the parameters $\mu$, $\nu$, and $\psi$ 
for the skill prior network $q_\psi(z_t|s_t)$, skill encoder $q_\mu(z_t|a_{t:t+H-1})$, and skill decoder $p_\nu(a_{t:t+H-1}|z_t)$ 
by minimizing Eq.(\ref{equation:joint_loss}). 
Adversarially, the discriminator  $\mathcal{D}_\zeta$ is updated by
minimizing Eq.(\ref{equation:pu_loss}),
which is equivalent to 
maximizing Eq.(\ref{equation:joint_loss}). 

%%%%%%%%%%%%%%%%%%%%%%%%%%%%%%%%%%%%%%%%%%%%%%%%%%%%%%%%%%%%%%%%%%%%%%%%%%%%%%%
\subsection{Downstream Policy Training} 
\label{section:dpt}
In the downstream policy training stage, we aim to exploit the skill knowledge learned
from the heterogeneous demonstration data, encoded by the skill prior network $q_\psi(\cdot)$ 
and decoder network $p_\nu(\cdot|z_t)$, to accelerate the online RL process.
To this end, we train a skill-based policy network $\pi_\theta(z_t|s_t)$ 
for the downstream online RL task with skill-level behavior cloning. 

Specifically, 
when interacting with the environment, 
we sample a skill $z_t$ from the skill policy network $\pi_\theta(\cdot|s_t)$ given the current state $s_t$.
The skill $z_t$ is then decoded to an action sequence $a_{t:t+H-1}$ using the 
skill decoder $p_\nu(\cdot|z_t)$ to guide the RL agent to reach state $s_{t+H}$ in $H$ steps.
The cumulative reward over the $H$ steps, i.e., the $H$-step reward, can be collected 
from the environment as $\tilde{r}_t=\sum_t^{t+H}r_t$.
With the online skill-based transition data $D=\{(s_t, z_t, \tilde{r}_t, s_{t+H})\}$,
we deploy a {\em Skill-based} soft actor-critic (SSAC) algorithm~\cite{pertsch2021accelerating}
to conduct skill-based policy learning with skill-level behavior cloning. 
SSAC extends soft actor-critic (SAC)~\cite{haarnoja2018soft} 
to learn the skill policy function network $\pi_\theta(z_t|s_t)$ (i.e., actor)
with the support of a skill-based soft Q-function network $Q_\phi(s_t,z_t)$ (i.e., critic). 
In particular, 
SSAC learns the skill policy function network 
by maximizing the following regularized expected skill-based Q-value:
\begin{align}
\label{equation:policy}
J_\pi(\theta)
	=\mathop{\mathbb{E}}_{\substack{s_t\sim D,\\ z_t\sim\pi_\theta}}
	\big[Q_\phi(s_t,z_t)
 -\kappa \mathcal{L}_{KL}(\pi_\theta(z_t|s_t),q_\psi(z_t|s_t))\big].
\end{align}
Different from the standard SAC algorithm, which regularizes the policy function
through its KL-divergence from a uniform distribution, 
the KL-divergence regularization term in SSAC
enforces the skill policy function $\pi_\theta(z_t|s_t)$
to clone the behavior of the pre-trained skill prior network $q_\psi(z_t|s_t)$.
From the value function's perspective, 
the KL-regularizer in Eq.(\ref{equation:policy}) 
can be interpreted as imposing a penalty on the skill-level Q-value
in the context of behavior cloning.
The soft skill-level Q-value function, $Q_\phi(s_t, z_t)$, 
is trained to minimize the following soft Bellman residual objective:
\begin{align}
\label{equation:critic}
J_Q(\phi)
	=\!\!\!\!\!\!\!
	\mathop{\mathbb{E}}_{(\substack{s_t,z_t,\tilde{r}_t,\\ s_{t+H}})\sim D}
	\left[\frac{1}{2}\big(Q_\phi(s_t,z_t)
    -(\tilde{r}_t \!+\! \gamma V_{\bar{\phi}}(s_{t+H}))\big)^2\right], 
\end{align}
which is computed on the data collected from the online interaction with the environment,
and $V$ denotes the state value function.  
To facilitate the learning of $Q_\phi$, 
we keep a stabilized Q-function $Q_{\bar{\phi}}(\cdot,\cdot)$ by computing its parameters
$\bar{\phi}$ from $\phi$ using an exponential moving average (EMA):
$\bar{\phi}\leftarrow\tau\phi+(1-\tau)\bar{\phi}$, where $\tau\in(0,1]$ is a momentum parameter. 
Based on \cite{levine2018reinforcement}, 
we then compute $V_{\bar{\phi}}(s_{t+H})$ from the penalized 
and stabilized soft Q-function $Q_{\bar{\phi}}(s_{t+H}, z_{t+H})$ as: 
\begin{align}
V_{\bar{\phi}}(s_{t+H})
& =\mathbb{E}_{z_{t+H}\sim\pi_\theta}\big[Q_{\bar{\phi}}(s_{t+H}, z_{t+H})\nonumber \\
	&-\kappa \mathcal{L}_{KL}(\pi_\theta(z_{t+H}|s_{t+H}),q_\psi(z_{t+H}|s_{t+H}))\big],
\end{align}
which incorporates the behavior cloning regularization into the Q-function learning as well,
efficiently utilizing the prior knowledge acquired from the demonstration data.
The main process of the 
SSAC algorithm for downstream policy training 
is outlined in Algorithm~\ref{algorithm:dpt}.

%%%%%%%%%%%%%%%%%%%%%%%%%%%%%%%%%%%%%%%%%%%%%%%%%%%%%%%%%%%%%%%%%%%%%%%%%%%%%%%
\begin{algorithm}[t]
\caption{Online Skill-Policy Training with SDE}
\label{algorithm:dpt}
\textbf{Input:} 
	Skill prior network $q_\psi(\cdot)$, decoder $p_\nu(\cdot)$, $\eta\in(0,1)$\\
\textbf{Initialize:}
	Replay buffer $D$, skill policy $\pi_\theta(\cdot)$, critics $Q_\phi$ and $Q_{\bar{\phi}}$\\
\textbf{Output:} Trained skill policy network $\pi_\theta(\cdot)$\\
\textbf{Procedure:}
\begin{algorithmic}[1]
\FOR{each iteration}
\FOR{every $H$ environment steps}
\STATE Sample skill $z_t$ from policy: $z_t\sim\pi_\theta(z_t|s_t)$\label{algorithm:sde}
\STATE Sample ${\bf a}_t$ = $a_{t:t+H-1}$ from decoder $p_\nu(\cdot|z_t)$
\STATE Sample state $s_{t+H}$ 
	and cumulative reward $\tilde{r}_t$ by interacting with environment using ${\bf a}_t$ 
	\STATE Update buffer: $D\leftarrow D\cup \{s_t, z_t, \tilde{r}_t, s_{t+H}\}$\\[.2em]
	\% {\em SDE augmentation steps in line 7-8:}\\[.2em] 
\STATE Sample $\epsilon\sim \mathcal{N}^m(0, 1)$, and let $\hat{z}_t=z_t+\eta\epsilon$ 
\STATE Augment buffer:  $D\leftarrow D\cup \{s_t, \hat{z}_t, \tilde{r}_t, s_{t+H}\}$
\ENDFOR
\FOR{each gradient step}
	\STATE Update policy parameters $\theta$ by maximizing Eq.(\ref{equation:policy})
	\STATE 
	Update Q-function parameters $\phi$ by minimizing Eq.(\ref{equation:critic})
	\STATE 
	EMA update $\bar{\phi}\leftarrow\tau\phi+(1-\tau)\bar{\phi}$
\ENDFOR
\ENDFOR
\end{algorithmic}
\end{algorithm}
%
%%%%%%%%%%%%%%%%%%%%%%%%%%%%%%%%%%%%%%%%%%%%%%%%%%%%%%%%%%%%%%%%%%%%%%%%%%%%%%%
\subsection{Skill-Level Data Enhancement}
\label{section:slda}
Collecting expert demonstrations can be challenging and expensive, due to the involvement of human experts~\cite{brys2015reinforcement}.
The scarcity of the limited expert data however hinders the robust and effective learning of skills. 
To alleviate the problem,
in addition to incorporating general, low-cost demonstration data during the prior training stage, 
we further propose a Skill-level Data Enhancement (SDE) technique to 
augment the skill-level data in 
both the skill prior training stage and the downstream policy training stage,
improving the robustness of learning at the skill-level.

Conventional data augmentation is typically applied to the input data, e.g., state observations.
It is not applicable to the action space since actions (continuous or discrete) cannot be easily re-represented or rescaled. 
By using a latent variable model to learn skills as 
latent representations for behaviours captured by action sequences, 
our proposed approach provides a {\em new augmentation space at the skill level},
without interfering with the real action space. 
Specifically, we propose to augment our skill level data 
with Gaussian noise altered versions as follows. 
For each skill embedding $z_t$, 
we can add a Gaussian noise altered version $\hat{z}_t=z_t+\eta\epsilon$ into 
the learning process,
where $\epsilon\sim \mathcal{N}^m(0, 1)$ is a Gaussian noise vector sampled from a $m$-dimensional
independent standard Gaussian distribution, $m$ is the dimension of the skill embedding,
and $\eta\in(0,1)$ is a very small scaling factor. 
We then enforce $z_t$ and $\hat{z}_t$ {\em correspond to the same action sequence},
aiming to achieve stable representations 
for different behaviors in the skill embedding space, 
and enhance the robustness of skill learning. 

In the skill prior training stage, 
SDE is realized by adding the following auxiliary reconstruction loss 
into the learning objective in Eq.(\ref{equation:joint_loss}):
\begin{align}
\!\!\!
\hat{L}_{rec}(\nu,\mu)=\alpha\!\!\!\!
	\!\!\mathop{\mathbb{E}}_{{\bf a}_{t}\sim A_{\pi_e}}\!\!\!
	\ell_{ls}\!\left(\hat{\bf a}_t\! \sim\! p_\nu\big(\cdot|\hat{z}_t\sim\eta\epsilon\!+\!
	q_\mu(\cdot|{\bf a}_t)\big),{\bf a}_t \right),
\end{align}
where $\alpha$ is a trade-off parameter. 
This augmenting loss adds Gaussian noise to the encoded skill vector, 
aiming to enforce the robustness of the skill encoder and decoder functions,
and make them resistant to minor variations in the skill embedding vectors. 

In the downstream skill policy training stage, SDE is realized efficiently by 
augmenting the skill-based transition data. 
For each observed skill-based transition $\{s_t, {z}_t, \tilde{r}_t, s_{t+H}\}$, 
we produce an altered skill vector $\hat{z}_t$ from $z_t$ and 
add an augmenting transition $\{s_t, \hat{z}_t, \tilde{r}_t, s_{t+H}\}$ to buffer $D$,
without any extra interaction with the environment. 
The goal is to make the skill-policy network more robust to small variations
in the skill embedding space, accelerating the learning process.  
The online skill policy training process augmented with SDE 
is outlined in Algorithm~\ref{algorithm:dpt}.
%

%%%%%%%%%%%%%%%%%%%%%%%%%%%%%%%%%%%%%%%%%%%%%%%%%%%%%%%%%%%%%%%%%%%%%%%%%%%%%%%
%
\section{Experiment}
\begin{figure}[t]
\centering
\setlength{\abovecaptionskip}{0.cm}
\includegraphics[width=0.40\linewidth,height=0.9in]{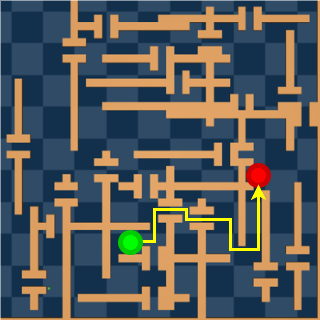}\hskip 0.15in
\includegraphics[width=0.40\linewidth,height=0.9in]{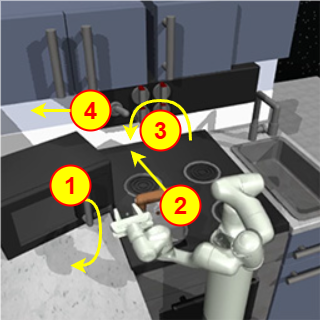} \vskip 0.15in
\includegraphics[width=0.40\linewidth,height=0.9in]{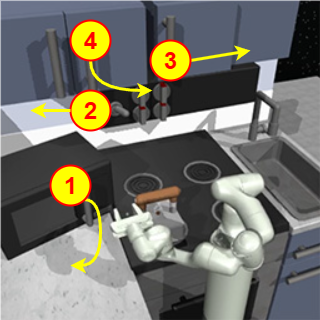}\hskip 0.15in
\includegraphics[width=0.40\linewidth,height=0.9in]{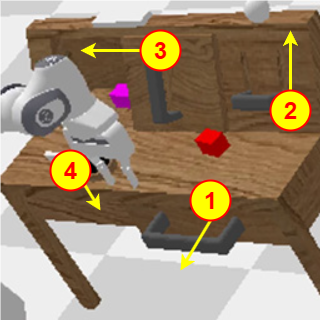}
\vskip .1in
\caption{The four environments used in the experiments. 
	Top row: \textbf{Maze} and \textbf{Kitchen}. 
	Bottow row: \textbf{Mis-aligned Kitchen} and \textbf{CALVIN}.
	In \textbf{Maze}, a point agent navigates from a starting point (green) to the target point (red).
	In each of the other three robotic manipulation environments, a robot arm 
	completes four different sub-tasks in order.
	}
\vskip 0.3in
\label{fig:envs}
\end{figure}
%%%%%%%%%%%
%%%%%%%%%%%
\begin{figure*}[t]
\centering
{\includegraphics[width=0.41\textwidth]{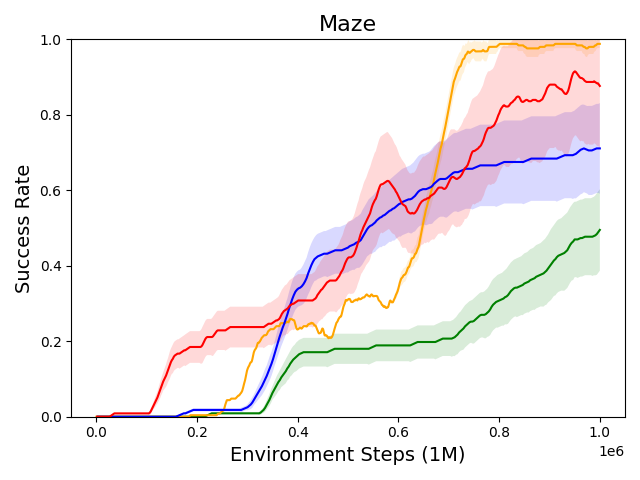}} \hskip .05in
{\includegraphics[width=0.41\textwidth]{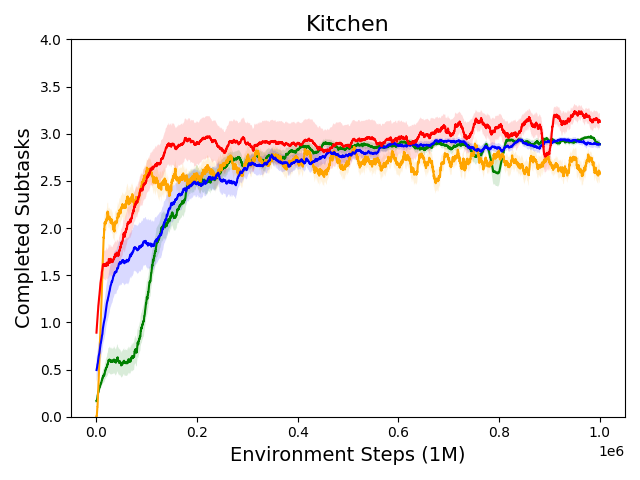}}\\
{\includegraphics[width=0.41\textwidth]{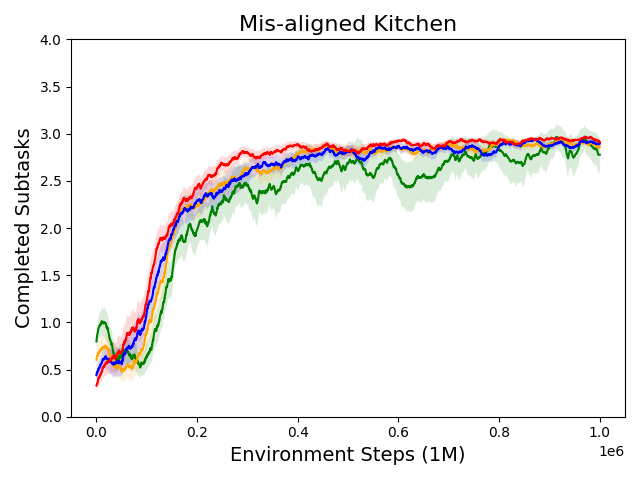}}\hskip .05in
{\includegraphics[width=0.41\textwidth]{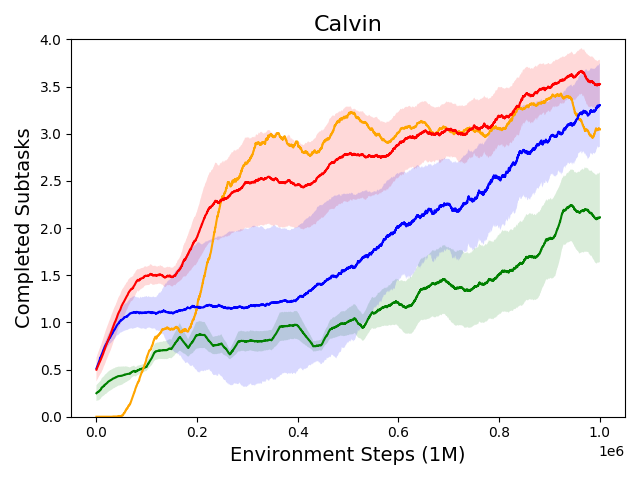}}\\
{\includegraphics[width=0.5\textwidth]{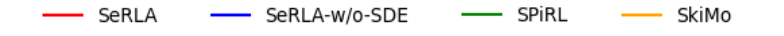}}
\vskip .05in
\caption{The comparison results on the four long-horizon sparse-reward tasks 
	(Maze, Kitchen, Mis-aligned Kitchen, and CALVIN) 
	are presented in the figure. 
	Each plot presents the average per-trajectory 
	return (i.e., reward) v.s. environment steps for each environment
	during the downstream training process. 
	The results were collected through 5 random seeds.}
\vskip 0.2in
\label{fig:experiment1}
\end{figure*}
%%%%%%%%%%%
\begin{figure}[t]
\centering
{\includegraphics[width=.41\textwidth]{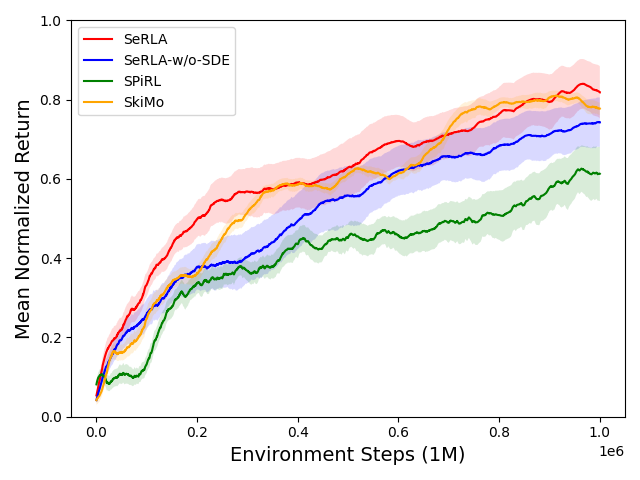}}
\vskip .05in
\caption{
	The plot summarizes the results across the four environments 
	and presents the mean normalized return averaged over the four environments v.s. environment steps.  
	The results were collected through 5 random seeds.}
\vskip 0.3in
\label{fig:experiment2}
\end{figure}
%%%%%%%%%%%

%%%%%%%%%%%%%%%%%%%%%%%%%%%%%%%%%%%%%%%%%%%%%%%%%%%%%%%%%%%%%%%%%%%%%%%%%%%%%%%
\subsection{Experimental Setting}

\paragraph{Environments}
We conduct experiments with four demonstration-guided tasks with long-horizon and sparse rewards 
in four different environments
that are commonly used for skill learning: Maze, Kitchen, Mis-aligned Kitchen, and CALVIN~\cite{shi2022skillbased}, which are shown 
in Figure~\ref{fig:envs}.  
The first three environments (Maze, Kitchen, and Mis-aligned Kitchen) are adapted from the D4RL datasets~\cite{fu2020d4rl}, 
while the last environment (CALVIN) is adapted from the CALVIN challenge~\cite{mees2022calvin}.
{\bf Maze} is a navigation environment, 
in which a point mass agent 
is required to find a path between a fixed starting point and a goal point. 
It is modified by randomly initializing the starting point of the agent around the original starting point~\cite{pertsch2021demonstrationguided}. 
The RL agent receives a sparse reward of 1 only when it reaches the close neighbor of the goal point
and receives a reward of 0 otherwise. 
{\bf Kitchen} is a robotic manipulation environment, 
in which a robotic arm completes a sequence of four sub-tasks  
(Microwave--Kettle--Bottom Burner--Light)~\cite{gupta2019relay}. 
The RL agent receives a sparse binary reward only when it completes each sub-task 
in sequence, and can obtain four reward scores if it completes all four sub-tasks in order.
{\bf Mis-aligned Kitchen} is modified from the Kitchen environment with a different task sequence
(Microwave--Light--Slide Cabinet--Hinge Cabinet)~\cite{pertsch2021demonstrationguided,shi2022skillbased}. 
Unlike in the original Kitchen setting, 
here the subtask order in expert demonstrations 
is misaligned with that in the downstream task, which makes skill learning more challenging.
{\bf CALVIN} is a robotic manipulation environment 
designed for Language-Conditioned Policy Learning~\cite{mees2022calvin}. 
The environment 
has been adapted for skill learning in SkiMo~\cite{shi2022skillbased}, 
requiring the RL agent to complete a sequence of 
four sub-tasks in order: Open Drawer, Turn on Lightbulb, Move Slider Left, and Turn on LED. 
The agent receives a binary reward signal for each sub-task it completes in order,
with the maximum value of per-trajectory reward as 4. 
%

%%%%%%%%%%%%%%%%%%%%%%%%%%%%%%%%%%%%%%%%%%%%%%%%%%%%%%%%%%%%%%%%%%%%%%%%%%%%%%%
\begin{table*}[t]
\centering
\caption{This table shows the percentage
improvements achieved by variants of SeRLA
	with Skill-level Data Enhancement (SDE) over the {\em ablation baseline SeRLA-w/o-SDE}. 
 The results are the average percentage increases of each method over SeRLA-w/o-SDE 
	across the whole downstream training stage.\\[.5em]
}
\label{table:ablation}
{
\begin{tabular}{l|cccc}
\Xhline{.5pt}
Average Increase & Maze & Kitchen & Mis-aligned Kitchen & CALVIN \\ \hline
SeRLA\_SDE (skill) & $0.22\pm0.30$ & $0.041\pm0.010$ & $0.032\pm0.025$ & $0.35\pm0.30$ \\
SeRLA\_SDE (downstream) & $0.15\pm0.19$ & $0.024\pm0.052$ & $0.021\pm0.018$ & $0.29\pm0.26$ \\
SeRLA (full) & $0.26\pm0.34$ & $0.087\pm0.066$ & $0.038\pm0.026$ & $0.41\pm0.35$ \\
\Xhline{.5pt}
\end{tabular}%
}
\vskip .1in
\end{table*}

%%%%%%%%%%%%%%%%%%%%%%%%%%%%%%%%%%%%%%%%%%%%%%%%%%%%%%%%%%%%%%%%%%%%%%%%%%%%%%%

%%%%%%%%%%%%%%%%%%%%%%%%%%%%%%%%%%%%%%%%%%%%%%%%%%%%%%%%%%%%%%%%%%%%%%%%%%%%%%%
\paragraph{Comparison Methods}
We compared the proposed SeRLA with two state-of-the-art skill-based methods, 
which train skill priors for the downstream tasks:  
\begin{itemize}
\item \textbf{SPiRL}~\cite{pertsch2021accelerating} learns a skill prior using a deep latent variable model from expert 
	demonstration data to guide policy training in the downstream task. 
\item \textbf{SkiMo}~\cite{shi2022skillbased} is a model-based method. It learns reusable skills and 
	a skill dynamic model in offline training, and selects the optimal skill in the downstream training using long-term model-based planning.
\end{itemize}

%%%%%%%%%%%%%%%%%%%%%%%%%%%%%%%%%%%%%%%%%%%%%%%%%%%%%%%%%%%%%%%%%%%%%%%%%%%%%%%
\paragraph{Implementation Details}
\label{sec:implementation}

We built SeRLA on top of SPiRL~\cite{pertsch2021accelerating} for skill prior training, and 
used the official implementations of the two comparison methods,  
SPiRL~\cite{pertsch2021accelerating} 
and SkiMo~\cite{shi2022skillbased}. 
The skill horizon was fixed to $H=10$ and the skill embedding dimension is set as $m=10$.
In skill-level adversarial PU learning, the discriminator $\mathcal{D}_\zeta$ is implemented as an MLP 
that has two 256-unit hidden layers with ReLU activations, followed by a Xavier-initialized linear output head. 
We set the positive class prior to $\lambda=0.5$, the relaxation slack variable to $\xi=0$, and the PU loss trade-off parameter to $\rho=0.1$. 
The trade-off parameter $\beta$ 
for the Gaussian prior regularizer is set to the same value as in SPiRL. 
For the SDE, the scaling factor is fixed at $\eta=0.01$, and 
the trade-off parameter for the augmenting loss is set as $\alpha=0.1$.
Downstream policy learning follows the SAC formulation~\cite{haarnoja2018soft} 
with $\tau=0.005$ and $\gamma=0.99$, while $\kappa$ is treated as a dual variable and updated during training.

\paragraph{Datasets}
The expert dataset $D_{\pi_e}$ contains demonstrations 
obtained either from human experts or from fully trained RL agents. 
We used the version aggregated in the SkiMo~\cite{shi2022skillbased} repository, 
which contains expert demonstrations collected in the four environments:
Maze, Kitchen, Mis-aligned Kitchen, and CALVIN. 
In {\em Maze}, the expert demonstration data is originally collected 
in the work of SPiRL~\cite{pertsch2021demonstrationguided}, 
consisting of 3,046 trajectories. 
In {\em Kitchen} and {\em Mis-aligned Kitchen}, 
the expert data is originally from the D4RL dataset~\cite{fu2020d4rl}, 
comprising 603 trajectories. 
In {\em CALVIN}, the expert data is from the CALVIN challenge~\cite{mees2022calvin}, comprising 1,239 trajectories.
In each environment, 
the general, low-cost demonstration data $D_\pi$ was collected using an RL agent 
that is pre-trained with $10^5$ timesteps of interaction with the environment, starting from scratch. 
In comparison to the full training regime of $10^7$ timesteps, 
the pre-trained RL agent is significantly undertrained, operating with a near-random policy. 
However, it has learned some basic behaviors that support exploration of the environments with minimal training cost. 
In each environment, 
the general low-cost demonstration data contains ten times as many trajectories as the expert data.

%%%%%%%%%%%%%%%%%%%%%%%%%%%%%%%%%%%%%%%%%%%%%%%%%%%%%%%%%%%%%%%%%%%%%%%%%%%%%%%
\subsection{Experimental Results}

We conducted experiments on the four environments (Maze, Kitchen, Mis-aligned Kitchen, and CALVIN)
to compare the proposed full method, SeRLA, and 
its variant without SDE, SeRLA-w/o-SDE, with the other two skill-based comparison methods, SPiRL and SkiMo. 
The skills were learned on the heterogeneous demonstration data prior to the downstream task 
and the reward was evaluated in the downstream RL learning process over $10^6$ environment steps. 
The maximum trajectory reward for the Maze environment is $1$, while for the Kitchen, Mis-aligned Kitchen, and CALVIN environments, it is $4$. 
The experimental results are presented in Figure~\ref{fig:experiment1} and Figure~\ref{fig:experiment2}. 
In Figure~\ref{fig:experiment1}, 
the four plots report results (return v.s. environment steps) on the four environments separately. 
We can see that the proposed SeRLA-w/o-SDE largely outperforms the baseline SPiRL across all the four environments,
especially on Maze, Mis-aligned Kitchen and CALVIN,  
which validates the effectiveness and contribution of the proposed PU Learning component 
in extracting useful skill knowledge from heterogeneous demonstration data. 
The proposed full approach SeRLA further improves performance over SeRLA-w/o-SDE,
yielding notable gains across all the four plots,
which verifies the effectiveness of the proposed skill-level data enhancement technique. 
SeRLA also outperforms the model-based state-of-the-art SkiMo and produces best results on Kitchen and Mis-aligned Kitchen. 
On the other two environments, Maze and CALVIN, SeRLA yields comparable overall performance to SkiMo throughout the downstream RL training, 
while producing the best results 
in the early training stage. 

To present a more illustrative overall comparison across the four environments, 
we evaluate the performance of each method by calculating its mean normalized return across all four environments, 
with reward from each environment being normalized to the range of $[0, 1]$ based on its maximum possible reward~\cite{cobbe2020leveraging}. 
The mean normalized return is obtained by taking the average normalized reward across the four environments, 
and plotted in Figure~\ref{fig:experiment2}. 
The results clearly show that SeRLA outperforms SkiMo in the early training stage, 
achieving the best overall performance among all the comparison methods. 
This validates the effectiveness of the proposed approach. 
%

%%%%%%%%%%%%%%%%%%%%%%%%%%%%%%%%%%%%%%%%%%%%%%%%%%%%%%%%%%%%%%%%%%%%%%%%%%%%%%%
\subsection{Ablation Study}
The previous comparison results have shown that the full approach SeRLA outperforms 
SeRLA-w/o-SDE and validated the contribution of the SDE technique.  
We further conducted an ablation study to assess the impact of the SDE 
technique at the two training stages separately. 
We experimented with two variants of SeRLA: 
(1) SeRLA\_SDE (skill), which denotes SeRLA with SDE applied only on the skill prior training; 
(2) SeRLA\_SDE (downstream), which denotes SeRLA with SDE applied only on the downstream policy training. 
We evaluate these two variants and the full approach against the 
baseline SeRLA-w/o-SDE that entirely drops SDE from both training stages, 
and record the percentage increase of their performance values over that of SeRLA-w/o-SDE. 

We collected the average increase of each method over the entire downstream training stage
and reported the results in Table~\ref{table:ablation}.
The results show that adding SDE to either training stage separately can produce performance gains within the proposed SeRLA framework. 
In comparison, SDE is {\em much more effective} when applied on the skill prior training stage, particularly on the Maze and CALVIN environments. 
This validates the efficacy of SDE in boosting the robustness of skill embedding learning,
which consequently improves the performance of downstream RL tasks. 
When applying SDE on both training stages, there are still some
marginal performance increases than applying it on each stage separately. 
These results again validate the contribution of SDE to the proposed approach. 

%%%%%%%%%%%%%%%%%%%%%%%%%%%%%%%%%%%%%%%%%%%%%%%%%%%%%%%%%%%%%%%%%%%%%%%%%%%%%%%
\section{Conclusion}
In this paper, we proposed a novel two-stage skill-level learning method SeRLA
to exploit heterogeneous offline demonstration data and accelerate downstream RL tasks. 
SeRLA deploys skill-level adversarial PU learning 
to learn reusable skills from both limited expert demonstration data and large low-cost demonstration data.
Then a skill-based soft actor-critic algorithm is deployed 
to utilize the learned skill prior knowledge and accelerate 
the online downstream RL through skill-based behavior cloning. 
The proposed approach conveniently provides a new augmentation space at the skill level
without interfering with the real action space,
which enables novel skill-level data enhancement (SDE) in both training stages. 
Our experimental results on four benchmark environments demonstrate 
that SeRLA outperforms two state-of-the-art skill learning
methods, SPiRL and SkiMo, especially in the early downstream training stage. 

%%%%%%%%%%%%%%%%%%%%%%%%%%%%%%%%%%%%%%%%%%%%%%%%%%%%%%%%
\bibliography{serla}

\end{document}